\documentclass[11pt]{article}
\usepackage{coling2020}
\usepackage{times}
\usepackage{url}
\usepackage{latexsym}
\usepackage{balance}
\usepackage{bm}
\usepackage{amssymb}
\usepackage{amsmath}
\usepackage{url}
\usepackage{makecell}
\usepackage{multirow}
\usepackage{subfig}
\usepackage{graphicx} 
\usepackage{color}
\usepackage{colortbl}
\usepackage{textcomp}
\usepackage{CJK}
\usepackage{enumitem}
\usepackage{booktabs}
\usepackage{microtype}
\usepackage{arydshln}

\definecolor{forestgreen}{rgb}{0.0, 0.50, 0.0}
\definecolor{goldenbrown}{RGB}{251,138,36}
\usepackage{wrapfig}
\usepackage{subfig}

\definecolor{zptu}{RGB}{18, 141, 21}

\definecolor{forestgreen}{rgb}{0.0, 0.50, 0.0}
\definecolor{goldenbrown}{rgb}{0.6, 0.4, 0.08}

\definecolor{g1}{RGB}{217,217,254}
\definecolor{g2}{RGB}{198,198,253}
\definecolor{g3}{RGB}{180,180,252}
\definecolor{g4}{RGB}{162,162,252}

\definecolor{r1}{RGB}{254,236,236}
\definecolor{r2}{RGB}{254,217,217}
\definecolor{r3}{RGB}{253,198,198}
\definecolor{r4}{RGB}{253,180,180}
\definecolor{r5}{RGB}{252,128,127}

\makeatletter
\g@addto@macro\normalsize{%
  \abovedisplayskip 4pt plus 2pt minus 3pt%
  \belowdisplayskip \abovedisplayskip
  \abovedisplayshortskip 4pt plus2pt  minus3pt%
  \belowdisplayshortskip 4pt plus2pt minus3pt%
}
\usepackage{arydshln}

\colingfinalcopy 

\title{Context-Aware Cross-Attention for Non-Autoregressive Translation}

\author{Liang Ding$^\dagger$\thanks{~~~Work done when interning at Tencent AI Lab.}~~~~~Longyue Wang$^\ddagger$~~~~~Di Wu$^\S$~~~~~Dacheng Tao$^{\dagger}$~~~~~Zhaopeng Tu$^\ddagger$
\\
  $^\dagger$The University of Sydney\\
  \tt ldin3097@uni.sydney.edu.au~~dacheng.tao@sydney.edu.au \\
  ~~~~~~~~~~~~~~~~$^\ddagger$Tencent AI Lab~~~~~~~~~~~~~~~~~~~~~~~~~~~~~~~~~~~~~$^\S$Peking University\\
  {\tt \{vinnylywang,zptu\}@tencent.com}~~~~~~~~~{\tt inbath@163.com}}

\date{}

\begin{document}
\maketitle

\begin{abstract}
  Non-autoregressive translation (NAT) significantly accelerates the inference process by predicting the entire target sequence. However, due to the lack of target dependency modelling in the decoder, the conditional generation process heavily depends on the cross-attention. In this paper, we reveal a localness perception problem in NAT cross-attention, for which it is difficult to adequately capture source context. To alleviate this problem, we propose to enhance signals of neighbour source tokens into conventional cross-attention. Experimental results on several representative datasets show that our approach can consistently improve translation quality over strong NAT baselines. Extensive analyses demonstrate that the enhanced cross-attention achieves better exploitation of source contexts by leveraging both local and global information. 
\end{abstract}


\section{Introduction}
\label{intro}

\blfootnote{
}

Different from autoregressive translation~\cite[AT]{rnnsearch,transformer} models that generate each target word conditioned on previously generated ones, non-autoregressive translation~\cite[NAT]{NAT} models break the autoregressive factorization and produce the target words in parallel.
Given a source sentence $\bf x$, the probability of generating its target sentence $\bf y$ with length $T$ is defined by NAT as:
$p({\bf y}|{\bf x})=p_L(T|{\bf x}; \theta) \prod_{t=1}^{T}p({\bf y}_t|{\bf x}; \theta)$, where $p_L(\cdot)$ is a separate conditional distribution to predict the length of target sequence. 
As NAT models can predict all tokens independently and simultaneously, recent works have fully investigated their superiority on decoding efficiency~\cite{lee2018deterministic,ghazvininejad2019mask,gu2019levenshtein,kasai2020parallel,sun2019fast,Shu2020LaNMT,Ran2019GuidingNN}. However, there still exists a gap between AT and NAT models in terms of effectiveness.

\begin{CJK}{UTF8}{gbsn}
In encoder-decoder frameworks, the cross-attention module dynamically selects relevant source-side information (key) given a target-side token (query)~\cite{yang2020sublayer,wang2020rethinking}. Through qualitative and quantitative analyses, we found that it is difficult for the NAT decoder to adequately capture the source context due to the lack of autoregressive factorization. As shown in Table~\ref{tab:case-study}, when translating the Chinese word ``交往'', the source context word ``女孩'' should play a significant role in predicting the candidate word ``dating''. However, the NAT model inappropriately generates ``socializing with'', resulting in lexical choice errors. As seen, the AT model gives relatively higher attention weights to local contexts on the source side while the NAT model pays less attention on them (0.15 vs 0.04). We make further statistical analysis in Section~\ref{sec:defination} to prove the universality of this localness perception problem.
\end{CJK}
Similar to our findings, \newcite{li2019hint} showed that distributions of cross-attention in NAT models are more ambiguous than those in AT ones.

To alleviate this localness perception problem in NAT, we propose a context-aware cross-attention to model both local and global contexts simultaneously. 
For local attention, we limit the scope of cross-attention to adjacent tokens surrounding the source word with the maximum alignment probability. We then combine the local attention weights with the original global ones by a gating mechanism (in Section~\ref{sec:method}).

Experiments are conducted on four commonly-cited datasets on translation task (i.e. WMT16 Romanian$\Rightarrow$English, WAT17 Japanese$\Rightarrow$English, WMT14 English$\Rightarrow$German and WMT17 Chinese$\Rightarrow$English) and show that our approach can consistently improve translation quality by around 0.5 BLEU point over advanced NAT models (in Section~\ref{sec:exp}). 
Further analyses reveal that our method can enhance abilities of NAT to learn syntactic and semantic information as well as phrase patterns (in Section~\ref{sec:analysis}).

\begin{CJK}{UTF8}{gbsn}
\begin{table}[tb]
\centering
\scalebox{1}{
\begin{tabular}{r|p{10cm}}
\hline
\textbf{Input} & \small 弗兰克 找到 一间 公寓 ， 同时 在 跟 一个 \textcolor{goldenbrown}{女孩} \textcolor{red}{\bf 交往} 。
\\
\hdashline
Reference & Frank found an apartment and was \textcolor{forestgreen}{\bf dating} a girl at the same time.\\
\hline
\textbf{NAT Output} & Frank found an apartment and was \textcolor{blue}{\em socializing with} a girl.\\
\rowcolor{g1}
{Attention} & \small 交往$_{0.68}$~~~弗兰克$_{0.18}$~~~\textcolor{goldenbrown}{女孩}$_{0.04}$\\
\hdashline
\textbf{AT Output} & Frank found an apartment and was \textcolor{blue}{\bf dating} a girl.\\
\rowcolor{g1}
{Attention} & \small 交往$_{0.81}$~~~\textcolor{goldenbrown}{女孩}$_{0.15}$~~~弗兰克$_{0.03}$\\
\hdashline
\textbf{Ours Output} &   Frank found an apartment and was \textcolor{blue}{\bf dating} a girl.\\
\rowcolor{g1}
{Attention} & \small 交往$_{0.69}$~~~\textcolor{goldenbrown}{女孩}$_{0.11}$~~~弗兰克$_{0.09}$\\
\hline
\end{tabular}
}
\caption{\label{tab:case-study}Case study of localness perception problem. ``NAT Output'' and ``AT Output'' are generated by NAT and AT models, respectively. ``Attention'' shows top-3 cross-attention probabilities when generating the target word ``dating'' or other equivalents. 
}
\vspace{-10pt}
\end{table}
\end{CJK}

\section{Localness Perception Problem}
\label{sec:defination}
\begin{CJK}{UTF8}{gbsn}
To validate our motivation, we conduct a statistical analysis. 
Following \newcite{tu2014localness}, we employ the locality entropy to measure how the cross-attention concentrate around a source word that corresponds with ${\bf y}_t$. 
As shown in Table~\ref{tab:case-study}, when generating the target side word ``dating'', the concentrated source word is ``交往'' according to the maximum probability of attention. 
And AT's attention distribution is obviously concentrated than NAT's, thereby have a lower entropy. In our case,
given a sentence pair $\{f_1,f_2,\dots,f_n; e_1,e_2,\dots,e_m\}$, for each decoding position $pos\in [1,m]$, we can obtain a probability distribution ${\bf P}_{pos}^i = \{P^i(f_1|{pos}), \dots, P^i(f_n|{pos})\}$ by calculating cross-attention in the $i$-th decoding layer. Thus, the locality entropy of one certain sentence is $\mathrm{LE}=-\frac{1}{6m}\sum_{i\in [1,6]}\sum_{pos\in [1,m]}{\bf P}_{pos}^i log_2{\bf P}_{pos}^i$. 
Finally, we average all sentence-level $\mathrm{LE}$ to get the corpus-level one. The lower LE means the more concentrated attention on source-side localness and vice versa. 
\end{CJK}

\begin{wraptable}{r}{7.3cm}
\centering
\vspace{-10pt}
    \begin{tabular}{l||cc|cc}
        \multirow{2}{*}{\bf Models} & \multicolumn{2}{c|}{\bf En-De} & \multicolumn{2}{c}{\bf Zh-En}\\
        \cline{2-5}
        &\textit{LE}&BLEU&\textit{LE}&BLEU\\
        \hline\hline
        NAT & 1.66 & 27.0 & 2.65 & 24.0\\
        ~~~+Ours & 1.62 & 27.5 & 2.51 & 24.6\\
        \hdashline
        AT & 1.46 & 29.2 & 2.12 & 25.3
    \end{tabular}
    \vspace{-5pt}
    \caption{The locality entropy \textit{LE} of NAT and AT models as well as our proposed method.}
    \vspace{-10pt}
    \label{tab:local-entropy}
\end{wraptable}

We compare the locality entropy of NAT and AT models on En-De and Zh-En. 
As shown in Table~\ref{tab:local-entropy}, the locality entropy ``\textit{LE}'' of NAT model is higher than that of AT, showing that the localness perception problem in NAT is more severe.
With the help of our method (in Section~\ref{sec:method}), this problem can be alleviated (\textit{LE}$\downarrow$), leading to better translation quality (BLEU $\uparrow$).  
This observation confirms the universality and side effect of localness perception problem in NAT, validating our hypothesis in Section~\ref{intro}.

\section{Context-Aware Cross-Attention for NAT}
\label{sec:method}

\begin{CJK}{UTF8}{gbsn}
\begin{figure}
    \centering
    \subfloat[Vanilla Non-autoregressive model]{\includegraphics[width=0.4\textwidth]{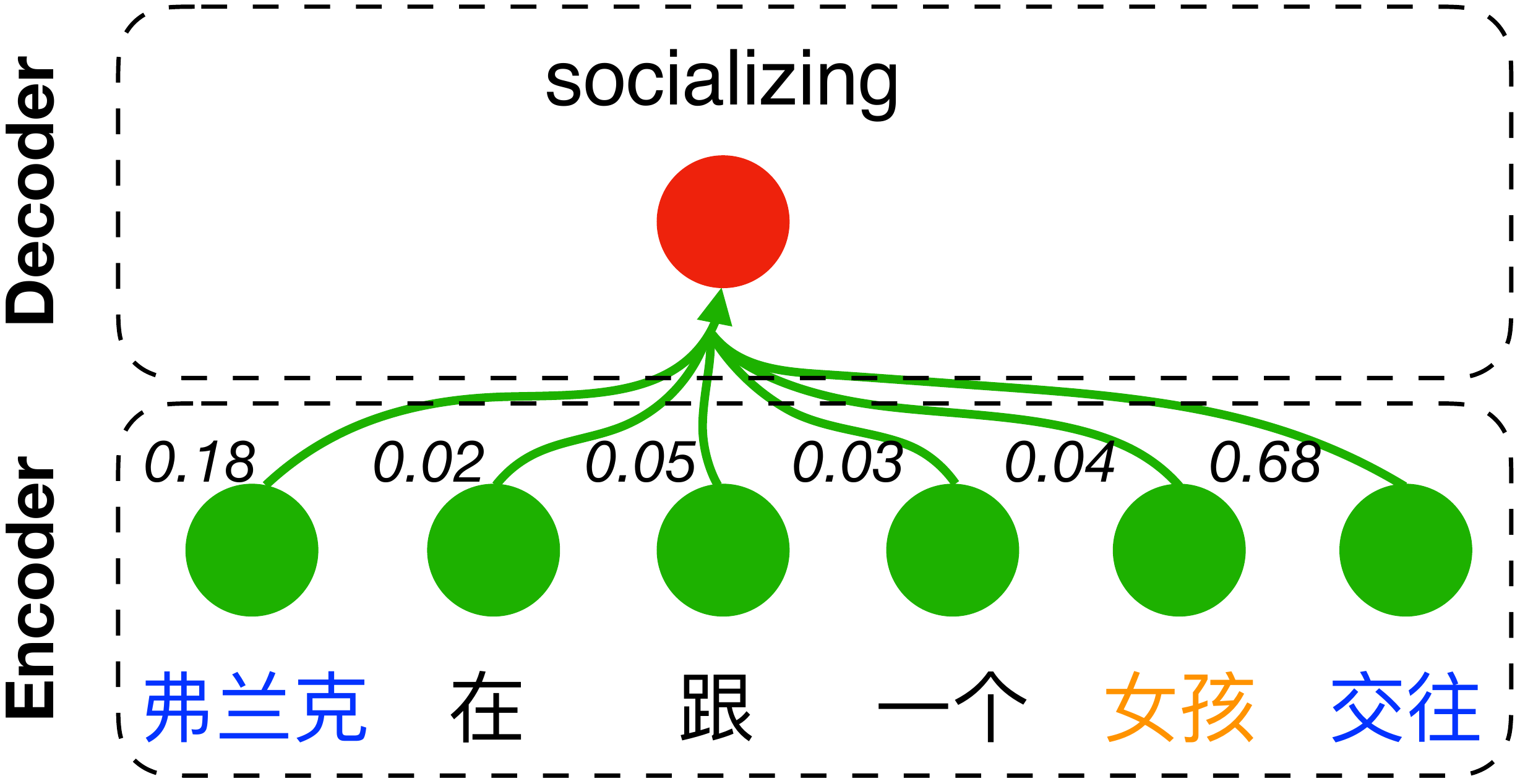}}
    \hspace{0.08\textwidth}
    \subfloat[Localness modeling in fixed window ]{\includegraphics[width=0.4\textwidth]{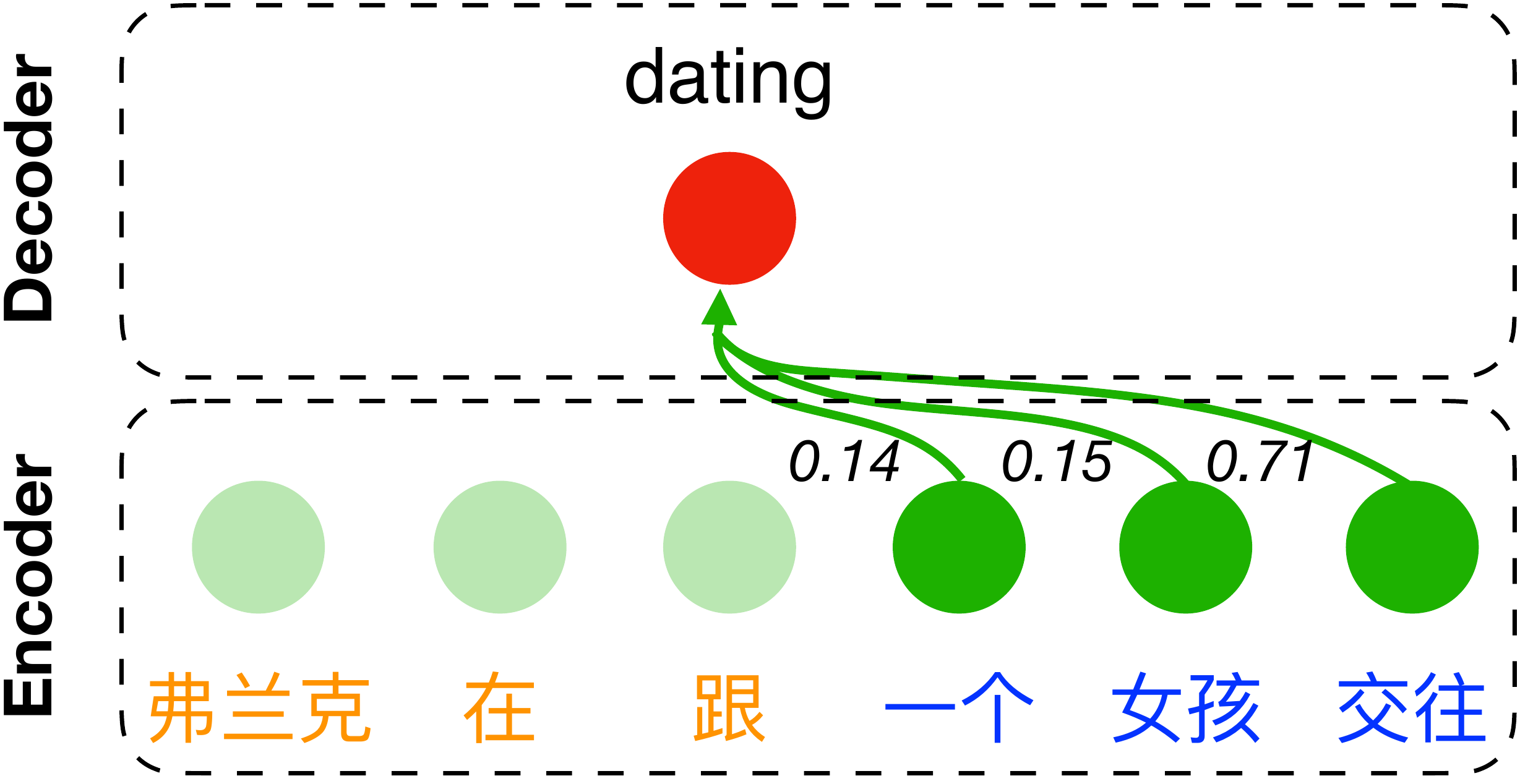}}
    \caption{Illustration of our proposed approach, which combines (a) vanilla cross-attention and (b) localness-aware cross-attention. In (a), the word ``交往'' is assigned with the maximum attention weight while the adjacent word (local context) ``女孩'' is assigned with a low weight. In (b), we guide the model to perceive the local context.}
    \label{fig:strcture}
    \vspace{-10pt}
\end{figure}
\end{CJK}

In this section, we introduce the detail of our proposed context-aware cross-attention networks (CCAN), which perceives the original and local cross-attention simultaneously.

\paragraph{Original Cross-Attention}
For the target-side query $Q$, source-side key $K$ and value $V$.
The $i$-th original cross-attention $\psi_{i}$ can be calculated with dot-product:
$\psi_{i}=Q_{i}K^{T}$. The original attention of the $i$-th element is the weighted sum of values $\textsc{Att}( \psi_{i}, V)=softmax(\psi_{i})V$  (in Figure~\ref{fig:strcture}(a)).

\paragraph{Our Approach} For the $i$-th position in target side, we propose a locally-sensitive cross-attention component for NAT to capture the neighbor signals. For simplicity, we adopt a straightforward but has been proven effective way~\cite{luong2015effective,xu-etal-2019-leveraging,you2020hard}: constricting the attention scope to a nearby window around the aligned $j$-th element. In practice, we choose the source element with the highest attention weight as the aligned element, and the local range can be modeled as follows:
\begin{equation}
L(\psi _{i})=\left\{
\begin{array}{rcl}
\psi _{i,j} & & {i-win\leqslant j \leqslant i+win}\\
-\infty & & {otherwise}
\end{array} \right.
\end{equation}
where $\psi_{i,j}$ denotes the attention correlation between the $i$- and $j$- elements in encoder and decoder parts, respectively. The $win$ is the hard-coded localness modeling window. Furthermore, we design an interpolation gating mechanism to wisely combine the original and local cross-attention:
\begin{equation}
\textrm{CCAN}(\underbrace{Q_i}_{\text{Decoder}},\underbrace{K,V}_{\text{Encoder}})=g\cdot \textsc{Att}(\psi _{i}, V) + (1-g)\cdot \textsc{Att}(L(\psi _{i}), V) \\
\label{eq:CCAN}
\end{equation}
where $g=\sigma (WQ_i)$ is the interpolation weight conditioned on the decoder side query $Q_i$ and $\sigma (\cdot )$ denotes the sigmoid function. Note that $W$ is the only additional parameter to estimate the importance of original cross-attention operation, and we share it for different cross-attention heads. 

\section{Experiments}
\label{sec:exp}

\subsection{Setup}

\paragraph{Data} Experiments are conducted on four widely-used translation datasets, including the \textit{small}-scale WMT16 Romanian-English (Ro-En,~\newcite{NAT}), the \textit{medium}-scale WMT14 English-German (En-De,~\newcite{transformer}), the \textit{large}-scale WMT17 Chinese-English (Zh-En,~\newcite{hassan2018achieving}), and word-order-divergent WAT17 Japanese-English (Ja-En,~\newcite{morishita2017ntt}), which consist of 0.6M, 4.5M, 20M and 2M sentence pairs, respectively. 
We preprocessed data via BPE~\cite{Sennrich:BPE} with 32K merge operations. We used BLEU~\cite{papineni2002bleu} as metric with statistical significance test~\cite{collins2005clause}.

\paragraph{Models} 
We follow \newcite{NAT} to apply sequence-level knowledge distillation~\cite{kim2016sequence} to simplify the training data. About AT Teachers, we train both \textsc{Base} and \textsc{Big} Transformer~\cite{transformer} models with corresponding training data. In \textsc{Big} model, we adopt large batch strategy (458K tokens per batch) to optimize the performance.
The main results employ Transformer-\textsc{Big} for all directions except Ro-En, which is distilled by \textsc{Base}. 
Our approach can be applied to different NAT architectures. In this paper, we mainly implement it on conditional masked language models~\cite[CMLMs]{ghazvininejad2019mask} and leave further investigation to future work. 
The model contains 6-layer encoder and 6-layer decoder, where the decoder trained with conditional mask language model fashion.
The model dimension is 512 on 8 heads, with 2048 feed forward dimensions. 
We follow the common practices~\cite{ghazvininejad2019mask,kasai2020parallel} to average the top three checkpoints to avoid stochasticity. 

\begin{table}[t]
    \centering
    \vspace{-1pt}
    \begin{tabular}{c|l||cc}
         \# &\bf Models &\bf BLEU &\bf $\Delta$\\
         \hline\hline
        1 & \textsc{base} & 26.5 & -- \\
         \hdashline
         2 & \textsc{ours} + win 3 & 26.8 & + 0.3\\
         3 & ~~~~~~~~~~ + win 5 & 26.8 & + 0.3\\
         4 & ~~~~~~~~~~ + win 7 & 26.7 & + 0.2\\
         5 & ~~~~~~~~~~ + win 9 & 26.9 & + 0.4\\
         6 & ~~~~~~~~~~ + win 11 & 26.7 & + 0.2\\
    \end{tabular}
    ~~~~~~~~~~
    \begin{tabular}{c|l||cc}
         \# &\bf Models &\bf BLEU &\bf $\Delta$\\
         \hline\hline
        1 & \textsc{base} & 26.5 & -- \\
         \hdashline
         2 & \textsc{win 9} + [1] & 26.6 & + 0.1 \\
         3 & ~~~~~~~~~~ + [1-3] & 26.7 & + 0.2 \\
         4 & ~~~~~~~~~~ + [6] & 26.8 & + 0.3 \\
         5 & ~~~~~~~~~~ + [4-6] & 26.8 & + 0.3\\
         6 & ~~~~~~~~~~ + [1-6] & 26.9 & + 0.4\\
    \end{tabular}
    \caption{Effects of localness range (left) and decoder layers (right) on translation quality.}
    \label{tab:ablation-window-layer}
\end{table}

\begin{table}[t]
    \centering
    \setlength{\tabcolsep}{3.2pt}
    \begin{tabular}{c|l|c||l|l|l|l}
    \#&\textbf{Models} &  {\bf Iteration} & {\bf Ro-En} & {\bf En-De} & {\bf Zh-En} & {\bf Ja-En}\\
    \hline
    \hline
    \multicolumn{7}{c}{\emph{Autoregressive}}\\
    \hline
    1&Transformer-\textsc{Base} 
    & n/a & 34.1  & 27.3 & 24.4 & 29.2 \\
    2&Transformer-\textsc{Big} 
    & n/a & n/a & 29.2 & 25.3 & 29.8 \\
    \hline \hline
    \multicolumn{7}{c}{\emph{Non-Autoregressive}}\\
    \hline
    3&NAT~\cite{NAT}    &   1   &   31.4 &19.2  &n/a & n/a\\
    4&Iterative NAT~\cite{lee2018deterministic} & 10  &30.2 &21.6 &n/a & n/a\\
    5&DisCo~\cite{kasai2020parallel}  &   4.8  &33.3& 26.8 &n/a & n/a \\
    6&Levenshtein~\cite{gu2019levenshtein} &  2.5  &33.3 &   27.3& n/a & n/a\\
    7&CMLMs~\cite{ghazvininejad2019mask} & 10  &  33.3 & 27.0 & 23.2 & n/a\\
    \hline
    \multicolumn{7}{c}{\emph{Our Implementation}}\\
    \hline
    8&CMLMs          &    \multirow{2}*{10}   &  33.3 & 27.0  & 24.0 & 28.9 \\
    9&~~~~+CCAN  &&   33.7 &  27.5$^\dagger$  & 24.6$^\dagger$ & 29.4$^\dagger$\\
    \end{tabular}
    \caption{Results of proposed method and comparison with previous work on WMT16 Ro-En, WMT14 En-De and WMT17 Zh-En datasets. ``$^\dagger$'' indicates statistically significant difference ($p < 0.05$) from the CMLM model.
    }
    \vspace{-10pt}
    \label{tab:main-results}
\end{table}

\subsection{Ablation Study}
In order to make best use of our proposed component for NAT, we conducted extensive ablation studies. All models are trained and validated on WMT14 En-De training and validation sets. 

\paragraph{Effects of Localness Range}
We investigate the localness window size within [3,5,7,9,11] and report the translation performance in Table~\ref{tab:ablation-window-layer} (left). As seen, our context-aware cross-attention with the window size of 9 achieves the best BLEU, which is therefore used as the default setting.

\paragraph{Effects of Decoder Layers}
As shown in Table~\ref{tab:ablation-window-layer} (right), deploying CCAN on the top-layer slightly outperforms deploying on the bottom-layer (``[6]''\textgreater``[1]''). In NAT, multiple decoding layers can be cast as the refiner, and the source central word chosen by the bottom-layer cross-attention is not as accurate as of the top-layer one. Our method, highly conditioned on the predicted central words, thus can gain a better effect on the top-layer compared to the bottom layer. 
In the end, modelling all layers (``[1-6]'') achieves the best performance and we thus use this setting in the following experiments.

\subsection{Main Results}

\begin{figure}[t]
    \centering
    \subfloat[Importance of localness on different layers.]{\includegraphics[width=0.382\textwidth]{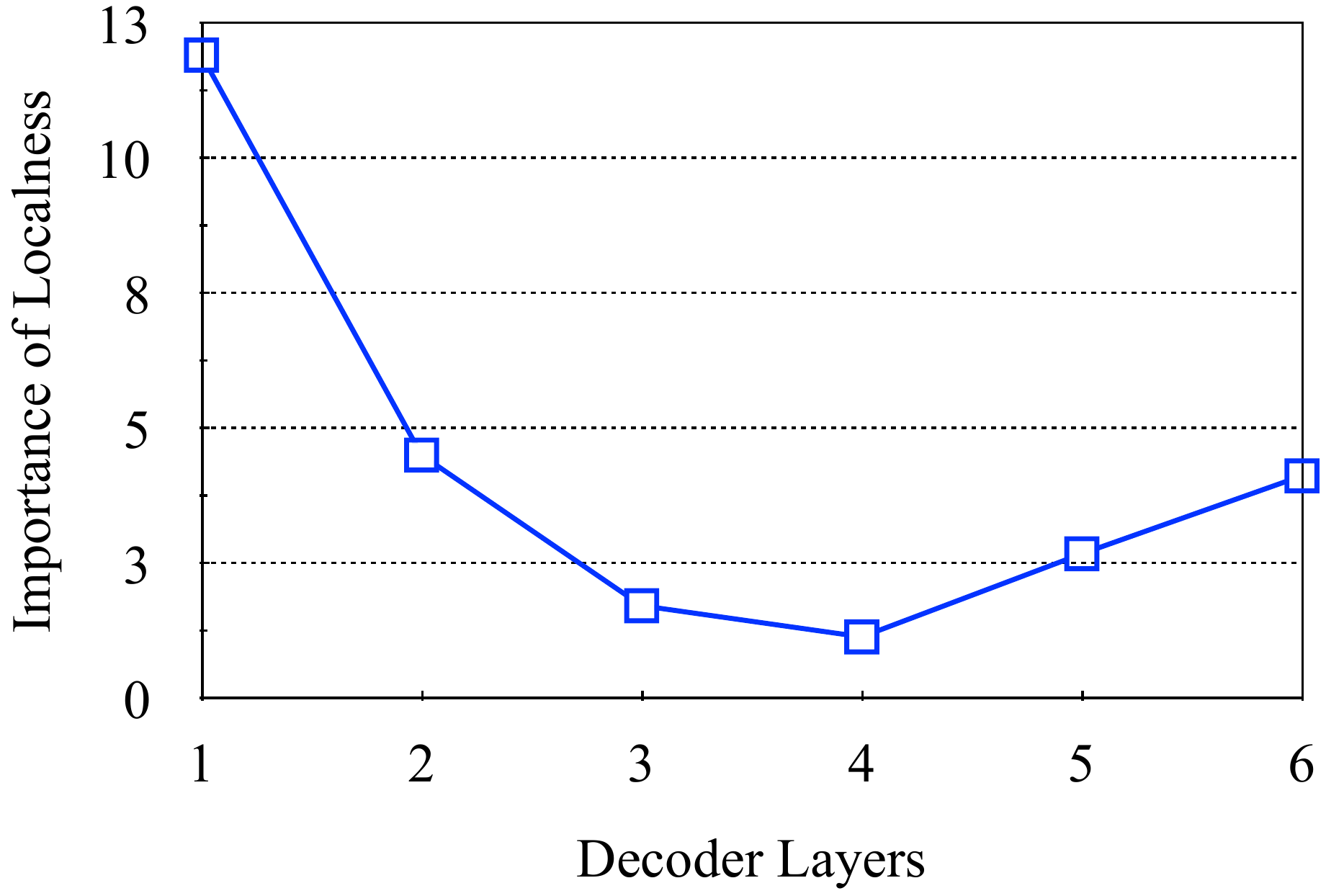}}
    \hspace{0.06\textwidth}
    \subfloat[Improvement on phrase translation.]{\includegraphics[width=0.4\textwidth]{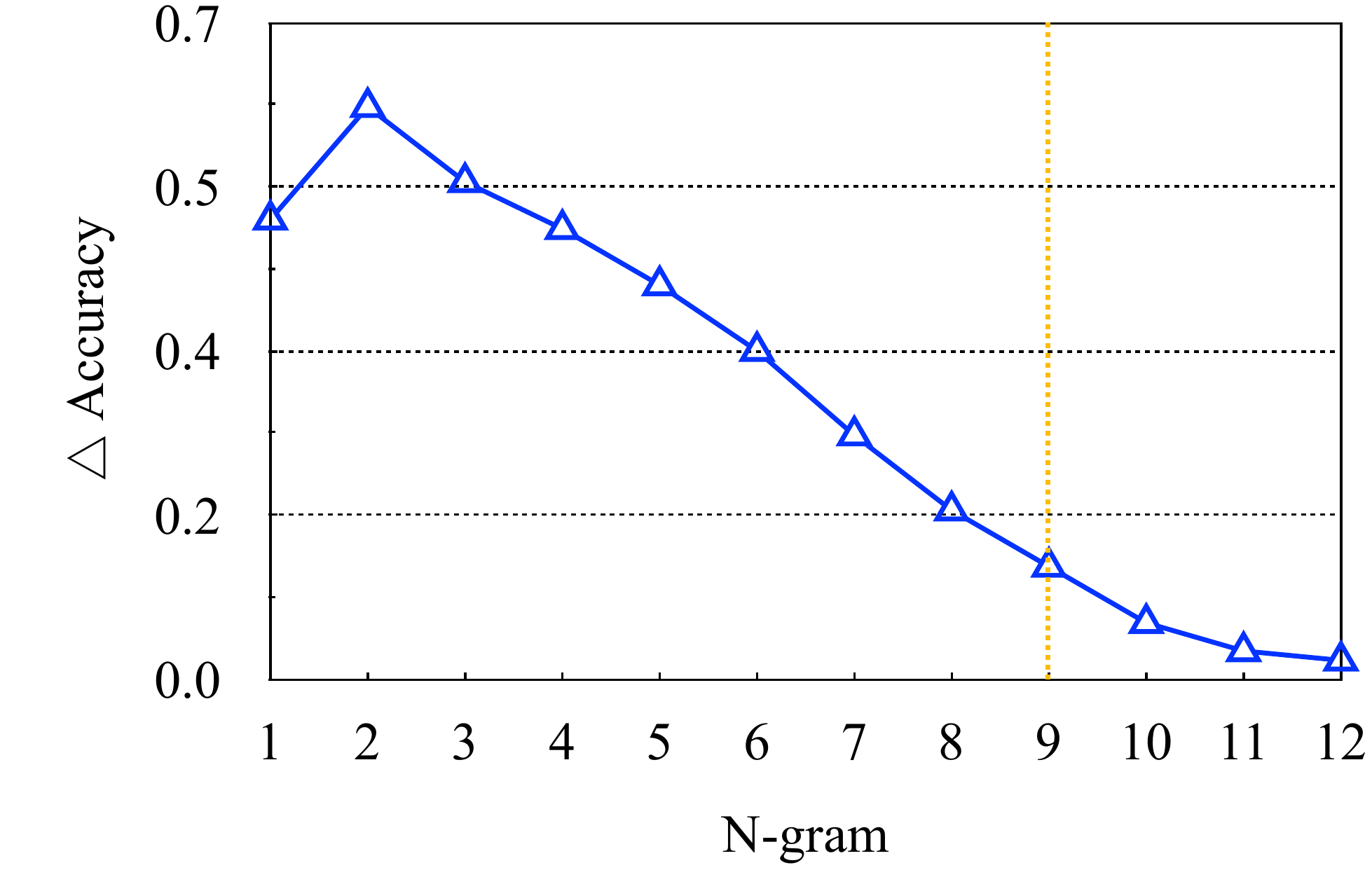}}
    \caption{Analyses on localness and phrasal patterns.}
    \label{fig:ablation}
\end{figure}

Table~\ref{tab:main-results} lists main results and comparison with previous NAT models on WMT16 Ro-En, WMT14 En-De, WMT17 Zh-En and WAT17 Ja-En datasets. We mainly implemented our approach on top of the advanced CMLMs model.
As seen, our approach (Row 9) consistently improves translation performance (BLEU$\uparrow$) over CMLMs on four language pairs. Note that our approaches only modify the cross-attention module and introduce fewer extra parameters, leading to negligible loss on latency. Encouragingly, our approach even slightly outperforms its AT teachers (Transformer-\textsc{Base}) on three tasks.

\section{Analysis}
\label{sec:analysis}
In this section, we conduct extensive analyses on WMT14 En-De to better understand how our method contribute to performance gains. 

\paragraph{Importance of Localness} 
The importance of localness should be different over layers. We explore it through gating (in Equation~\ref{eq:CCAN}) analyzing. Specifically, 
we cast the weighting scalar of local cross-attention as its importance degree and calculate the importance of localness for each decoder layer. As shown in Figure~\ref{fig:ablation}(a), during information flow evolving from bottom to top layers, the importance of localness continues to decline till the penultimate layer, and then increases. 
The possible reason for the increase in the last two layers is that the top layer followed by softmax, requiring more source-side context to choose lexicons. 

\paragraph{Phrasal Patterns} 

Our approach is expected to pay more attention to the most relevant source token and its neighbours, such that the phrasal translation can be improved. To evaluate the accuracy of phrase translations, we calculate the improvement on n-gram tokens in Figure~\ref{fig:ablation}(b), where the golden dashed line indicates that the window size is 9. As seen, CCAN consistently outperforms the baseline ($\Delta$Accuracy\textgreater0), indicating that our method can enhance the ability of NAT model on capturing the phrasal information, which is similar with~\newcite{yang2018modeling}'s findings.

\begin{table}[t]
    \centering
    \begin{tabular}{l||cc|ccc|ccccc}
    \multirowcell{2}{\textbf{Model}}&\multicolumn{2}{c|}{\textbf{Surface}}&\multicolumn{3}{c|}{\textbf{Syntactic}}&\multicolumn{5}{c}{\textbf{Semantic}}\\
    \cline{2-11}
    &SeLen&WC&TrDep&ToCo&BShif&Tense&SubNm&ObjNm&SoMo&CoIn\\
    \hline
    \hline
    CMLMs&\textbf{93.2}&\textbf{79.4}&45.8&79.4&73.5&\textbf{88.9}&87.5&85.4&52.7&\textbf{63.1}\\
    \hline
    ~~+CCAN&92.8&78.1&\textbf{46.5}&\textbf{79.7}&\textbf{74.1}&88.3&\textbf{88.1}&\textbf{85.7}&\textbf{52.9}&62.5\\
    \end{tabular}
    \caption{Results of probing tasks. We evaluate linguistic properties learned by sentence encoder. }
    \label{tab:probing}
\end{table}

\paragraph{Linguistic Properties}

Intuitively, our proposed cross-attention component brings context-aware representation, may affecting the linguistic properties learned by the encoder. 
We quantitatively investigate it from linguistic perspectives with probing tasks~\cite{conneau-etal-2018-cram}. These tasks can be categorized into three types: ``\textbf{Surface}'' focuses on the simple surface properties learned from the sentence embedding; ``\textbf{Syntactic}'' quantifies the syntactic reservation ability; and ``\textbf{Semantic}'' assesses the deeper semantic representation ability. 
To evaluate the representation ability of CCAN equipped NAT model, we compare the pre-trained vanilla NAT and CCAN equipped NAT encoders, followed by a MLP classifier. Specifically, the mean of the top encoding layer, as sentence representation, will be passed to the classifier. We can see from Table~\ref{tab:probing}, the CCAN equipped NAT encoder preserves rich syntactic and semantic information. 

\section{Conclusion and Future Work}
We reveal a localness perception problem in NAT. To alleviate it, we propose the context-aware approach to make the cross-attention pay more attention to source-side local words, which in turn improves the translation performance over several benchmarks.
In future work, we will investigate selectively choosing the context~\cite{geng2020does,yang2019context} rather than the fixed window size. Besides, it is interesting to enhance NAT model with extra signals, such as cross-lingual position embedding~\cite{ding2020self}, larger context~\cite{wang-cross2017} and pre-trained initialization~\cite{liu2020multilingual}.

\section*{Acknowledgements}

This work was supported by Australian Research Council Projects under grants FL-170100117, DP-180103424, and IC-190100031. We are grateful to the anonymous reviewers and the area chair for their insightful comments and suggestions.

\bibliographystyle{coling}
\bibliography{coling2020}

\end{document}